\documentclass{article}

\usepackage{times}
\usepackage{graphicx, epsfig} 
\usepackage{subfigure} 

\usepackage{natbib}

\usepackage{algorithm}
\usepackage{algorithmic}
\usepackage{amsmath}

\usepackage[accepted]{icml2014arxiv}

\usepackage{xspace, amsmath}

\providecommand{\mypara}[1]{\medskip\par \noindent \textsc{#1.}}

\newcommand{\OurModel}{{\textsc{EdgeExplain}}\xspace}
\newcommand{\Pone}{{\bf (P1)}\xspace}
\newcommand{\Ptwo}{{\bf (P2)}\xspace}
\newcommand{\Pthree}{{\bf (P3)}\xspace}
\newcommand{\Labelset}{\ensuremath{\mathcal{T}}\xspace}

\newcommand{\Lt}{\ensuremath{L(t)}\xspace}

\newcommand{\softmax}{\ensuremath{\mbox{softmax}}\xspace}

\icmltitlerunning{Joint Inference of Multiple Label Types in Large Networks}

\begin{document} 

\twocolumn[
\icmltitle{Joint Inference of Multiple Label Types in Large Networks}

\icmlauthor{Deepayan Chakrabarti}{deepay@fb.com}
\icmladdress{Facebook Inc.}
\icmlauthor{Stanislav Funiak}{sfuniak@fb.com}
\icmladdress{Facebook Inc.}
\icmlauthor{Jonathan Chang}{jonchang@fb.com}
\icmladdress{Facebook Inc.}
\icmlauthor{Sofus A. Macskassy}{sofmac@fb.com}
\icmladdress{Facebook Inc.}

\icmlkeywords{social networks, label inference}

\vskip 0.3in
]

\begin{abstract} 
We tackle the problem of inferring node labels in a partially labeled
graph where each node in the graph has multiple label {\em types} and
each label type has a large number of possible labels.  Our primary
example, and the focus of this paper, is the joint inference of label
types such as hometown, current city, and employers, for users
connected by a social network.  Standard label propagation fails to
consider the properties of the label types and the interactions
between them.  Our proposed method, called \OurModel, explicitly
models these, while still enabling scalable inference under a
distributed message-passing architecture.
On a billion-node subset of the Facebook social network,
\OurModel significantly outperforms label propagation for several
label types, with lifts of up to $120\%$ for recall@1 and $60\%$ for
recall@3.
\end{abstract}

\section{Introduction}
\label{sec:intro}

Inferring labels of nodes in networks is a common classification
problem across a wide variety of domains ranging from social networks
to bibliographic networks to biological networks and more. The
typical goal is to predict a single label of low dimensionality for
each node in the network (say, whether a webpage in a {\tt .edu}
domain belongs to a professor, student, or the department) given a
partially labeled network and possibly attributes of the nodes. In this paper
we instead consider the problem of inferring multiple fields such as
the hometowns, current cities, and employers of users of a social
network, where users often only partially fill in their profile, if at
all. Here, we have multiple {\em types} of missing labels, where each
label type can be very high-dimensional and correlated.
Joint inference of such label types is important for many ranking and relevance
applications such as friend recommendation, ads and content
targeting, and user-initiated searches for friends,
motivating our focus on this problem.

\begin{figure}[tp]
\begin{center}
\centerline{\epsfig{file=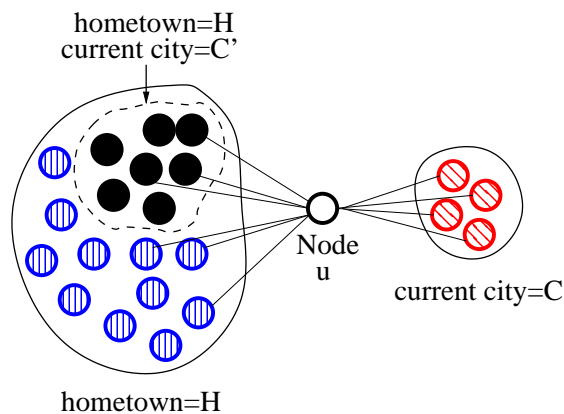,width=0.9\columnwidth}}
\caption{{\em An example graph of $u$ and her friends:}
The hometown friends of $u$ coincidentally contain a
subset with current city $C'$. This swamps the group from $u$'s
actual current city $C$, causing label propagation to infer $C'$ for $u$.
However, our proposed model (called \OurModel) correctly explains all
friendships by setting the hometown to be $H$ and current city to be
$C$.}
\label{fig:example}
\end{center}
\vspace{-1em}
\end{figure} 



One standard method of label inference is label
propagation~\cite{zhu02learning, zhu03semi}, which tries to set the
label probabilities of nodes so that friends have similar
probabilities. 
While this method
succinctly captures the essence of homophily (the more two nodes have
in common, the more likely they are to connect~\cite{McPherson2001}),
it optimizes for only a single type of
label and assumes only a single category of relationships.
It therefore fails
to address the potential complexity of 
edge formation in
networks,
where
nodes
have different reasons to link to each other. As an example,
consider the snapshot of a social network in 
Figure~\ref{fig:example}, where we want to predict the hometown 
and current city of node $u$, given what we know about $u$ and $u$'s
neighbors. Here, the labels of node $u$ are
completely unknown, but her friends' labels are completely known.
Label propagation would treat each label independently and infer the
hometown of $u$ to be the most common hometown among her friends, the
current city to be the most common current city among friends, and so
on.  Hence, if the bulk of friends of $u$ are from her hometown $H$,
then inferences for current city will be dominated by the most common
current city among her hometown friends (say, $C'$) and not friends
from her actual current city $C$; indeed, the same will happen for all
other label types as well.

Our proposed method, named \OurModel, approaches the problem from a
different viewpoint, using the following intuition: Two nodes 
form an edge for a {\em reason} that is likely to be related to
them sharing the value of one or more label types (e.g., two users went
to the same college).  Using this intuition, we can go beyond standard
label propagation in the following way: instead of taking the graph as
given, and modeling labels as items that propagate over this graph, we
consider the labels as factors that can {\em explain} the observed
graph structure.  For example, the inferences for $u$ made by label
propagation leave $u$'s
edges from $C$ completely
unexplained. Our proposed method rectifies this, by trying to infer
node labels such that for each 
edge $u\sim v$, we can point to a reason why this is so --- $u$ and
$v$ are friends from the same hometown, or college, or the like. While
we are primarily interested in inferring labels, we note that the
inferred reason for each
edge
can be important applications by
itself; e.g., if a new 
node $u$ joins a
network and forms
and edge with $v$, knowledge of the reason can
help with
the well-known link prediction task --- should we recommend $v$'s
college friends, or friends from the same high school, etc.?

We note that a seemingly simple alternative solution --- cluster the
graph and then propagate the most common labels within a cluster ---
is in fact quite problematic.  In addition to the computation cost,
any clustering based solely on the graph structure ignores labels
already available from user profiles, but any clustering that tries to
use these labels must deal with incomplete and missing labels. The
clustering must also be complex enough to allow many overlapping
clusters. Hence, we believe that clustering does not readily lend
itself to a solution for our problem.




\newpage
Our contributions are:

\begin{enumerate}
\item We formulate the label inference problem as one of explaining
  the graph structure using the labels. We explicitly account for the
  fact that labels belong to a limited set of label {\em types}, whose
  properties we enumerate and incorporate into our model.
\item 
  Our gradient-based iterative method for inferring labels is easily
  implemented in large-scale message-passing architectures. We
  empirically demonstrate its scalability on a billion-node subset of
  the Facebook social network, using publicly available user profiles
  and friendships.
\item 
  On this large real-world dataset,
  \OurModel
  significantly outperforms label propagation for several label types,
  with lifts of up to $120\%$ for recall@1 and $60\%$ for recall@3.
  These improvements in accuracy, combined with the scalability of
  \OurModel, clearly demonstrate its usefulness for label
  inference on large networks.

\end{enumerate}

The paper is organized as follows. We survey related work in
Section~\ref{sec:related}. Our proposed model is discussed in
Section~\ref{sec:proposed}, followed by the inference method in
Section~\ref{sec:inference}, and generalizations of the model in
Section~\ref{sec:general}. Empirical evidence proving the
effectiveness of our method is presented in Section~\ref{sec:exp},
followed by conclusions in Section~\ref{sec:conc}.


\section{Related Work}
\label{sec:related}
We discuss prior work in semi-supervised learning, statistical
relational learning, and in latent models for networks.

\smallskip
\noindent
\textsc{Semi-supervised learning.}
Many graph-based approaches can be viewed as estimating a function
over the nodes of the graph, with the function being close to the
observed labels, and smooth (similar) at adjacent nodes.
Label propagation~\cite{zhu03semi,zhu02learning} uses a quadratic
function, but other penalties are also
possible~\cite{zhou04learning,belkin04regularization,belkin05on}.  Other
approaches modify the random walk interpretation of label
propagation~\cite{baluja08video,talukdar09new}.
In order to handle a large number of distinct label values, the label
assignments can be summarized using count-min
sketches~\cite{talukdar14scaling}. None of the
approaches consider interactions between multiple label types, and
hence fail to capture the edge formation process in graphs considered
here.

\smallskip
\noindent
\textsc{Statistical relational learning.}
These algorithms typically predict a label based on
(a) a local classifier that uses a node's
attributes alone, 
(b) a relational classifier that 
uses the labels at adjacent nodes, and 
(c) a
collective inference procedure that propagates the information through
the network~\cite{chakrabarti98enhanced, 
perlich03aggregation, lu03link, macskassy07classification}.
Macskassy et al.~\citeyearpar{macskassy07classification}
observe that the best algorithms 
(weighted-vote relational neighbor 
classifier~\cite{macskassy07classification}
with relaxation labeling~\cite{rosenfeld76scene,hummel83on}) tend to
perform as well as label propagation, which we outperform.  While
there has been some work focusing on understanding how to combine and
weigh different edge types for best prediction
performance~\cite{Macskassy2007}, the edge types (analogous to our reason for an edge) were
given up front.  We note that these algorithms typically focus on a
single label type, while we explicitly model the interactions among
multiple types.

There is also extensive work on probabilistic relational models,
including Relational Bayesian
Networks~\cite{koller98probabilistic,friedman99learning}, Relational
Dependency Networks~\cite{neville07relational}, and Relational Markov
Networks~\cite{taskar02discriminative}. These are very general
formalisms, but it is our explicit modeling assumptions regarding
multiple label types that yields gains in accuracy.

\smallskip
\noindent
\textsc{Latent models.}
Graph structure has been modeled using latent variables
\cite{hoff02latent,miller09nonparametric,palla12infinite}, but with an emphasis on
link prediction.  However, our goal is to make predictions about each
individual user, and such latent features can be arbitrary
combinations of user attributes, rather than concrete label types we
wish to predict.  Other models simultaneously explain the connections
between documents as well as their word
distributions~\cite{nallapati08joint,chang10hierarchical,ho12document}.  While we
do not consider the problem of modeling text data, our model permits
us to incorporate node attributes, such as group memberships.
Finally, the number of distinct label values in our application is
very large (on the order of millions), and we suspect that the
latent variables would have to have a large dimension to explain
the edges in our graph well.

\section{Proposed Model}
\label{sec:proposed}

%

In this section we first build intuition about our model using a
running example.  Suppose we want to infer the {\em labels} (e.g.,
``Palo Alto High School'' and ``Stanford University'') corresponding
to several {\em label types} (e.g., high school and college) for a
large collection of users.  The available data consist of labels
publicly declared by some users, and the (public) social network among
users, as defined by their friendship network.  While the desired set
of label types may depend on the application, here we focus on five
label types: hometown, high school, college, current city, and
employer.

Our solution exploits three properties of these label types:
\begin{itemize}
\item[\Pone] They represent the primary situations where two people
     can meet and become friends, for example, because they went to
     the same high school or college.
\item[\Ptwo] These situations are (mostly) mutually exclusive.  
     While there may be occasional friendships sharing, say, hometown
	 and high-school, we make the simplifying assumption that most
	 edges can be explained by only one label type.
\item[\Pthree] Sharing the same label 
     is a {\em necessary} but not {\em sufficient} condition.
     For example, ``We are friends from Chicago''
     typically implies that the indicated individuals were, at some
     point in time, co-located in a small area within Chicago (say,
     lived in the same building, met in the same cafe), but hardly
     implies that two randomly chosen individuals from Chicago are
     likely to be friends.
\end{itemize}
\Pone is a direct result of our application; our desired label types
were targeted at friendship formation. 
Combined with
\Ptwo, our five label types can be considered a set of mutually
exclusive and exhaustive ``reasons'' for friendship; while this is
not strictly true for high school and hometown, empirical
evidence suggests that it is a good approximation (shown later in
Section~\ref{sec:exp}) and we defer a
discussion on this point to Section~\ref{sec:general}.
However, as \Pthree shows, we cannot simply
cast the labels as features whose mere presence or absence significantly
affects the probability of friendship; instead, a more careful
analysis is called for.

Formally, we are given a graph, 
$\mathcal{G}=(V,E)$ and a set of label types $\Labelset=\{t_1,\ldots,t_k\}$.
For each label type $t$, let $L(t)$ denote the (high-dimensional) set of labels
for that label type.  Each node in the graph is associated with binary
variables $S_{ut\ell}$, where $S_{ut\ell}=1$ if node $u\in V$ has label
$\ell$ for label type $t$.  Let $\boldsymbol S_V$ and $\boldsymbol S_H$
represent the sets of visible and hidden variables, respectively.  We want to
infer the correct values of $\boldsymbol S_H$, leveraging $\boldsymbol S_V$ and
$\mathcal{G}$.

A popular method for label inference is {\em label
  propagation}~\cite{zhu02learning,zhu03semi}.  For a single label
type, this approach represents the labeling by a set of indicator
variables $S_{u\ell}$, where $S_{u\ell}=1$ if node $u$ is labeled as
$\ell$ and 0 otherwise.  \citet{zhu03semi} relax the labeling to
real-valued variables $f_{u\ell}$ over all nodes $u$ and labels $\ell$
that are clamped to one (or zero) for nodes known to possess that
label (or not).  They then define a quadratic energy function that
assigns lower energy states to configurations where $f$ at adjacent
nodes are similar:
\begin{equation}
E(\boldsymbol f) = \frac{1}{2} \sum_{u\sim v} w_{uv} \sum_\ell (f_{u\ell} -
f_{v\ell})^2.
\label{eq:zhuEnergy}
\end{equation}
Here, $u\sim v$ means that $u$ and $v$ are linked by an edge, and
$w_{uv}$ is a non-negative weight on the edge $u\sim v$. The minimum
of Eq.~\ref{eq:zhuEnergy} is found by solving the fixed point
equations
\begin{equation}
f_{u\ell} = \frac{1}{d_u} \sum_{u\sim v} w_{uv} f_{v\ell},
\label{eq:zhuUpdate}
\end{equation}
where $d_u = \sum_{u \sim v} w_{uv}$. This procedure encourages
$f_{u\ell}$ of nodes connected to clamped nodes to be close to the
clamped value and propagates the labels outwards to the rest of the
graph.  Multiple label types can be handled similarly by minimizing
Eq.~\ref{eq:zhuEnergy} independently for each type. 



While this formulation
makes full use of \Pone and has the advantage of simplicity, it
completely ignores \Ptwo. Intuitively, label propagation
assumes that friends tend to be similar in {\em all} respects (i.e.,
all label types), whereas what \Ptwo suggests is that each friendship
tends to have a single {\em reason}: an edge $u\sim v$ exists
because $u$ and $v$ share the same high school {\em or} college {\em
or} current city, etc. This highly non-linear function is not easily
expressed as a quadratic or similar variant.

Instead, we propose a different probabilistic model, which we call
\OurModel.  As described above, let $\boldsymbol S_V$ and
$\boldsymbol S_H$ represent the sets of visible and hidden variables
respectively; the variable $S_{ut\ell}$ is known (visible) if user $u$
has publicly declared the label $\ell$ for type $t$, and unknown
(hidden) otherwise.  Then, \OurModel is defined as follows:
\begin{align}
P(\boldsymbol S_V, \boldsymbol S_H) &= \dfrac{1}{Z} \prod_{u\sim v} \underset{t\in\Labelset}{\softmax} (r(u, v, t))\label{eq:opt}\\
r(u, v, t) & = \sum_{\ell\in\Lt} S_{ut\ell} S_{vt\ell} \label{eq:optReason}\\
\underset{t\in\Labelset}{\softmax}(r(u, v, t)) &=
\sigma\biggl(\alpha\sum_{t\in\Labelset} r(u, v, t) +
c\biggr),\label{eq:optSoftmax}
\end{align}
where $Z$ is a normalization constant.
Here, $r(u, v, t)$ indicates whether a shared label type
$t$ is the {\em reason} underlying the edge $u\sim v$
(Eq.~\ref{eq:optReason}).
The $\softmax(r_1, \ldots, r_{|\Labelset|})$ function should have
three properties: (a) it should be monotonically non-decreasing in each
argument, (b) it should achieve a value close to its maximum as long
as any one of its parameters is ``high'', and also (c) it should be
differentiable, for ease of analysis. In Eq.~\ref{eq:optSoftmax}, 
we use the sigmoid
function to implement this: $\sigma(x) = 1/(1+e^{-x})$. This
monotonically increases from $0$ to $1$, and 
achieves values greater than $1-\epsilon$ once $x$ is greater than an
$\epsilon$-dependent threshold.
In addition, the sigmoid enables fine control of the degree of
``explanation'' required for each edge (discussed below) and
allows for easy extensions to more complex label types and extra
features (Section~\ref{sec:general}), all of which make it our
preferred choice for the \softmax.


In a nutshell, our modeling assumption can be stated as follows: {\em
It is better to explain as many friendships as possible, rather than
to explain a few friendships really well.}
Eq.~\ref{eq:opt} is maximized if the \softmax function
achieves a high value for each edge $u\sim v$, i.e., if each edge is
``explained''. This is achieved if the sum 
$\sum_{t\in\Labelset} r(u, v, t)$ is more than the required
threshold, which in turn is satisfied if the product
$S_{ut\ell}S_{vt\ell}$ is 1 for even one label $\ell$ --- in other words,
when there exists any label $\ell$ that both $u$ and $v$ share.
The parameter $\alpha$ controls the degree of explanation needed for
each edge; a small $\alpha$ forces the learning algorithm to be
very sure that $u$ and $v$ share one or more label types, while with a large
$\alpha$, a single matching label type is enough. Empirical results shown
later in Section~\ref{sec:exp} prove that large $\alpha$ values
perform better, suggesting that even a single matching label type is enough to
explain the edge. 
The parameter $c$ in Eq.~\ref{eq:optSoftmax} can be thought of as the
probability of matching on an unknown label type, distinct from the
five we consider. Higher values of $c$ can be used to model
uncertainty that the available label types form an {\em exhaustive}
set of reasons for friendships. All our experiments use $c=0$, and
reflect our belief in property \Pone; the accuracy of
predictions under this setting (shown later in Section~\ref{sec:exp})
suggests that \Pone indeed holds true.

Further intuition can be gained by considering a node $u$ whose labels are
completely unknown, but whose friends' labels are completely known
(see Figure~\ref{fig:example}).
As we discussed earlier in Section~\ref{sec:intro},
label propagation would infer the
hometown of $u$ to be the most common hometown among her friends
(i.e., $H$), the
current city to be the most common current city among friends (i.e.,
$C'$), and so on. 
However, such an
inference leaves $u$'s friendships from $C$
completely unexplained. Our proposed method rectifies this;
Eq.~\ref{eq:opt} will be maximized by correctly inferring $H$ and $C$
as $u$'s hometown and current city respectively, since $H$ is enough to
explain all friendships with the hometown friends, and the marginal
extra benefit obtained from explaining these same friendships a little
better by using $C'$ as $u$'s current city is outweighed by the
significant benefits obtained from explaining all the friendships from
$C$ by setting $u$'s current city to be $C$.

To summarize, Eq.~\ref{eq:optReason} encapsulates property \Pone by
trying to have matching labels between friends;
Eq.~\ref{eq:optSoftmax} models property \Ptwo by enabling
any one label type to explain each friendship; and the form of the 
probability distribution (Eq.~\ref{eq:opt}) uses only existing edges
$u\sim v$ and not all node pairs, and thus is not affected when, say,
two nodes with Chicago as their current city are not friends, which
reflects the idea that matching label types are necessary but not
sufficient \Pthree.



\section{Inference}
\label{sec:inference}

The probabilistic description of \OurModel in
Eqs.~\ref{eq:opt}-\ref{eq:optSoftmax} can be restated as an
optimization problem in the variables $S_{ut\ell}\in\{0, 1\}$. In the
spirit of~\cite{zhu03semi}, we propose a relaxation in terms of a
real-valued function $\boldsymbol f$, with $f_{ut\ell}\in[0,1]$ representing the
probability that $S_{ut\ell}=1$, i.e., the probability that user $u$ has label $\ell$ for
label type $t$. This yields the following optimization:
\begin{align}
\mbox{Maximize} & \sum_{u\sim v} \log\Bigl(
\underset{t\in\Labelset}{\softmax} (r(u, v, t))\Bigr) \label{eq:optRelaxed}\\
\mbox{where}\,\, r(u, v, t) & = \sum_{\ell\in\Lt} f_{ut\ell} f_{vt\ell}\\
\sum_{\ell\in\Lt} f_{ut\ell} & = 1 \quad\forall t\in\mathcal{T}\\
f_{ut\ell} & \geq 0 \label{eq:optRelaxedConstraints}
\end{align}
where $\softmax(.)$ is defined as in Eq.~\ref{eq:optSoftmax}, and the
equation for $r(.)$ is analogous to Eq.~\ref{eq:optReason} but
measures the total probability that $u$ and $v$ have the same
label for a given label type $t$.

The problem is not convex in $\boldsymbol f$, but is convex in
$\boldsymbol f_u = \{f_{ut\ell} | t\in\mathcal{T}, \ell\in\Lt\}$ if
the distributions $\boldsymbol f_v$ are held fixed for all nodes
$v\neq u$. Hence, we propose an iterative algorithm to infer
$\boldsymbol f$. Given $\boldsymbol f_v$ for all $v\neq u$,
finding the optimal $\boldsymbol f_u$ corresponds to solving the following
problem:
$$\mbox{Maximize } g(\boldsymbol f_u) = 
\sum_{v\in\Gamma(u)} \log\Bigl(\underset{t\in\Labelset}{\softmax} (r(u, v, t))\Bigr),$$
where the summation is only over the set $\Gamma(u)$ of the friends of
$u$, and we again restrict $\boldsymbol f_u$ to be a set of
$|\mathcal{T}|$ probability distributions, one for each label type.
We note that $g(.)$ is convex and Lipschitz continuous with constant
$L=\alpha\cdot|\Gamma(u)|$, where $|\Gamma(u)|$ is
the number of friends of $u$.

This is a constrained maximization problem with no closed form
solution for $\boldsymbol f_u$. 
To solve it, we use proximal gradient ascent, which is an iterative
method where in each step, we take a step in the
direction of the gradient, and then project it back to the
probability simplex $\Delta = \left\{f_{ut\ell}\mid f_{ut\ell}\geq
0, \sum_{\ell\in\Lt} f_{ut\ell}=1 \forall t\in\mathcal{T}\right\}$.
Specifically, let $\nabla g$ represent the gradient of $g$, with
components given by:
$$\dfrac{\partial g({\boldsymbol f_u})}{\partial f_{ut\ell}} =
\sum_{v\in\Gamma(u)}
\alpha f_{vt\ell} \cdot \sigma\biggl(-\alpha
\sum_{t\in\mathcal{T}}\sum_{\ell\in\Lt}f_{ut\ell}f_{vt\ell} - c\biggr).$$
Let $\boldsymbol f^{(k-1)}_u = \bigl\{f^{(k-1)}_{ut\ell} |
t\in\mathcal{T}, \ell\in\Lt\bigr\}$ be the estimated probability
distributions for each of the $\mathcal{T}$ label types
at the end of iteration $k-1$, and let
$q^{(k)}_{ut\ell}$ represent the (possibly improper)
ending point of the $k$-th gradient
step: 
$$\boldsymbol q^{(k)}_u = \boldsymbol f^{(k-1)}_u + c_k \nabla g,$$
where $c_k$ is a
step-size parameter that we could set to a constant $c_k = 1/L$.
The point $\boldsymbol q^{(k)}_u$ is now projected to the closest point in
$\Delta$: 
$$\boldsymbol f^{(k)}_u = \underset{\boldsymbol
q'\in\Delta}{\arg\min} \|\boldsymbol q^{(k)}_u - \boldsymbol q'\|_2.$$ 
This
can be easily achieved in expected linear time over the size of the
label set $\sum_t\Lt$~\cite{duchi08efficient}. If only sparse
distributions can be stored for each label type (say, only the top $k$
labels for each type), the optimal $k$-sparse projections
can be obtained simply by setting to $0$ all but the top $k$ labels
for each label type, and then projecting on to the simplex~\cite{kyrillidis13sparse}.

This algorithm converges to a fixed point, and the function values
converge to the optimal at a $1/k$ rate~\cite{beck09gradient}:
$$g^* - g^{(k)} \leq \dfrac{L\|\boldsymbol f^{(0)}_u - \boldsymbol
f^*_u\|^2}{2k} \leq \dfrac{L|\mathcal{T}|}{k},$$
where $\boldsymbol f^*_u$ represents the optimal set of probability distributions,
and $g^*$ is the optimal function value. 
An important consequence of the algorithm is that computation of $\boldsymbol f_u$
only requires information
from $\boldsymbol f_v$ for the neighbors $v$ of $u$. Thus, it is a ``local''
algorithm that can be easily implemented in distributed
message-passing architectures, such as Giraph~\cite{giraph,scaling}.

\section{Generalizations}
\label{sec:general}


We now discuss some aspects of \OurModel and some generalizations that
demonstrate its wide applicability.

\mypara{Related label types} 
Property \Ptwo assumes that the reasons for friendship formation are
mutually exclusive, but this need not be strictly true.
For example, some high school friends could be a subset of hometown
friends\footnote{The relationship between high school and hometown is
in fact more complicated. The high school could be within driving
distance of the hometown, but not in it; and sometimes even this does
not hold.}.  
Let us again consider Figure~\ref{fig:example}, but with
current city replaced by high school. Suppose that the solid-black
nodes represent actual high school friends, and we are trying to infer
$u$'s high school.  
If the small cluster on the right did not exist, then
Eq.~\ref{eq:opt} would be maximized by picking
the most common high school among $u$'s friends (i.e., the solid-black
nodes), even if they are
already explained by a shared hometown; thus, \OurModel would pick the
correct high school.  On the other hand, if some friendships would
remain unexplained without a shared high school (e.g., the small 
cluster in Figure~\ref{fig:example}), then it is not obvious whether
we should prefer a high school that explains these edges or a
high school that represents a large segment of hometown friends.
The
parameter $\alpha$ modulates this trade-off, with a higher value of
$\alpha$ emphasizing the explanation of all edges as against the
explanation of several edges a little better.
The choice of $\alpha$
must depend on the characteristics of the social network; for the
Facebook network, the best empirical results are achieved for
large $\alpha$ (shown later in Section~\ref{sec:exp}), suggesting that
many of our label types are indeed mutually exclusive.

\mypara{Incorporating user features}
\OurModel easily generalizes to broader settings with multiple user
features, such as group memberships, topics of interest, keywords, or pages
liked by the user. As an example, consider group memberships of users.
Intuitively, if most members of a group come from the same college,
then it is likely a college-friends group, and this can aid inference
for group members whose college is unknown. This can be easily handled
by creating a special node for each group, and creating
``friendship'' edges between the group node and its members. \OurModel
will infer labels for the group node as well, and will explain its
``friendships'' via the college label. This, in turn, will influence
college inference for group members with unknown college labels. The
importance of such group membership features can also be tuned, as
described next.

\mypara{Incorporating edge features}
There are several situations where edge-specific features could be
useful. First, we may want to give more importance to certain kinds
of edges, such as the group-membership edges mentioned above.
Second, some features could be important for one label type but not
another: e.g., the age difference between friends could be useful for
inferring high school but not employer. All these situations can be
easily handled by modifying Eq.~\ref{eq:optReason} to include an
edge-specific and label type-specific weight. The corresponding
modifications to the inference method are trivial.

\section{Experiments}
\label{sec:exp}


Previously, we provided intuition and examples
supporting the claim that \OurModel is better suited to inference
of our desired label types than vanilla label propagation. In this
section, we
demonstrate this via empirical evidence on a billion-node graph.

\mypara{Data}
We ran experiments on a large subgraph of the Facebook social
network, consisting of over $1.1$ billion
users and their
friendship edges.  From the public profile of each user, we extract
the hometown, current city, high school, college, and employer,
whenever these are available.  The dimensionality of our five label
types range from almost $900K$ to over $6.3M$.  We describe below in
{\sc Implementation Details} our process for generating the edges.
This forms our base dataset.

\begin{figure*}[ht]
\begin{center}
\begin{tabular}{cc}
\hspace{-1em}
\includegraphics[width=0.475\textwidth]{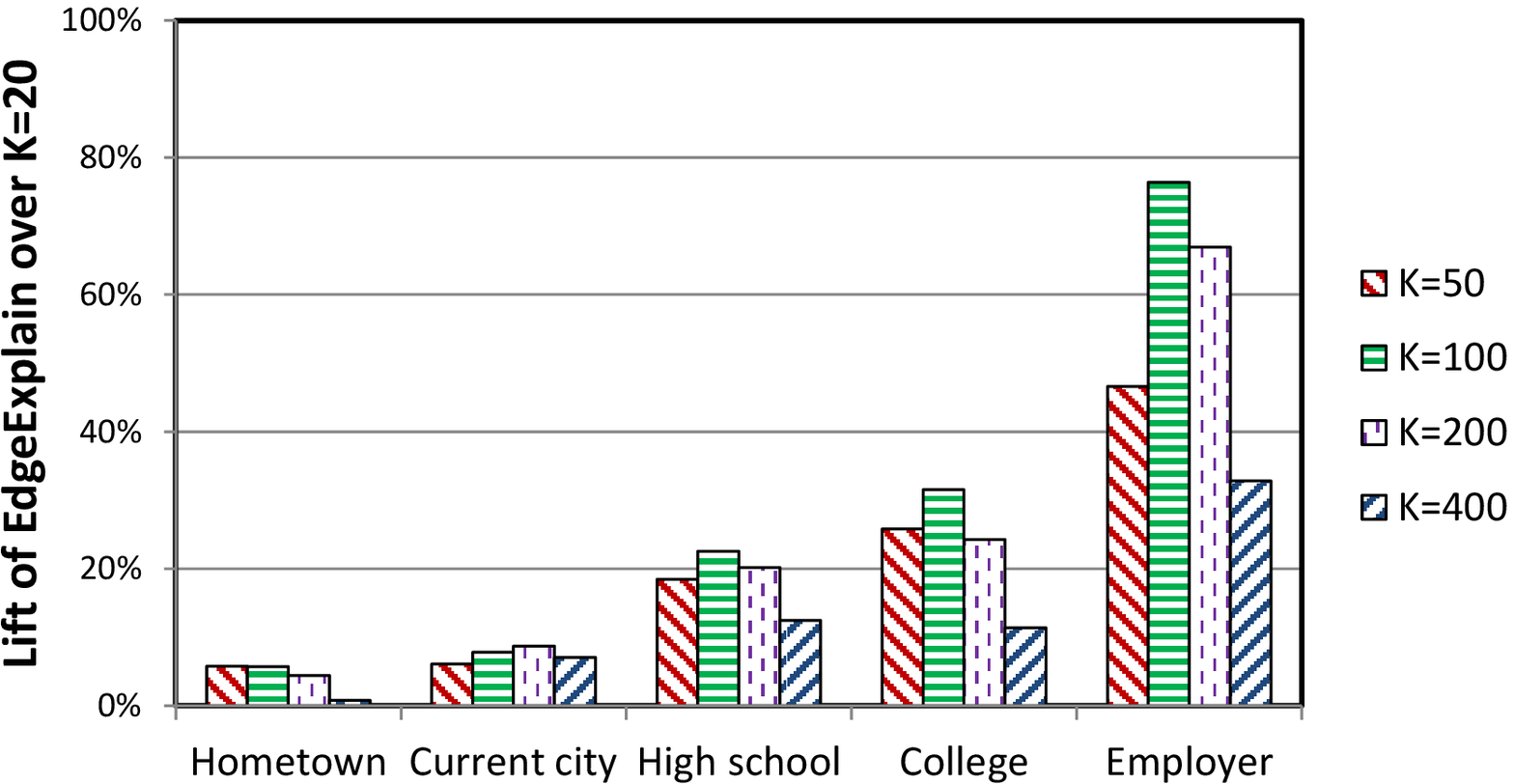} &
\includegraphics[width=0.475\textwidth]{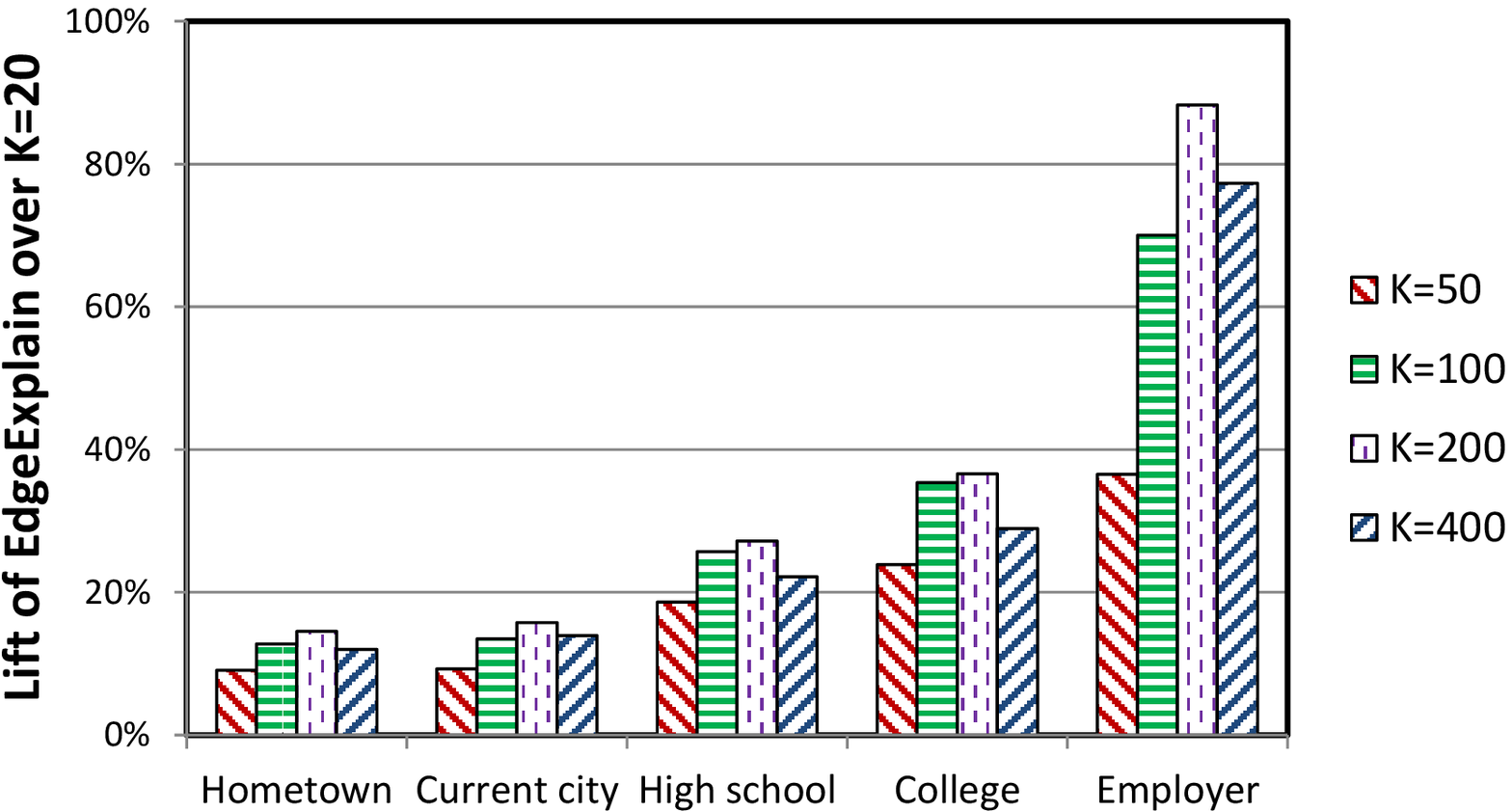}\\
\hspace{-1em}
(a) Recall at $1$ & (b) Recall at $3$
\end{tabular}
\caption{{\em Recall of \OurModel for graphs built with different
    number of friends $K$:} The plot shows lift in recall with respect
  to a fixed baseline of \OurModel with $K=20$.  Increasing $K$
  increases recall up to a point, but then the extra friends introduce
  noise which hurts accuracy.}
\label{fig:liftWRTbaseline}
\end{center}
\vspace{-0.5em}
\end{figure*}


\mypara{Experimental Methodology}
The set of users is randomly split into five parts and experimental
accuracy is measured via 5-fold cross-validation, with the known
profile information from four folds being used to predict labels for
all types for users in the fifth fold. 
Results over the various folds are identical to three decimal places.
All differences are therefore significant and we do not show variances
as they are too small to be noticeable.

In each experiment, we run inference on the training set and compute a
ranking of labels for each label type for each user. This ranking is
provided by $\boldsymbol f$ computed for label propagation
(Eq.~\ref{eq:zhuEnergy}) and \OurModel
(Eqs.~\ref{eq:optRelaxed}-\ref{eq:optRelaxedConstraints}) respectively.
We then measure recall at the top-$1$ and top-$3$ positions, i.e., we
measure the fraction of (user, label type) pairs in the test set
where the predicted top-ranked label (or any of the top-$3$ labels) match
the actual user-provided label. 
For reasons of confidentiality, we only present the {\em lift} in
recall values of \OurModel as compared to label propagation.

\mypara{Implementation Details}
We implemented \OurModel in Giraph~\cite{giraph,scaling}
which is an iterative graph
processing system based on the Bulk Synchronous Processing
model~\cite{malewicz10pregel, valiant90bridging}.
The entire set of nodes is split among $200$
machines, and in each iteration, every node $u$ sends the probability
distributions $\boldsymbol f_u$ to all friends of $u$. To limit the
communication overhead, we implemented two features: (a) for each user
$u$ and label type $t$, the multinomial distribution $f_{ut.}$ was
clipped to retain only the top $8$ entries
optimally~\cite{kyrillidis13sparse}, and (b) the friendship
graph is sparsified so as to retain, for each user $u$, the top $K$
friends whose ages are closest to that of $u$. This choice of friends
is guided by the intuition that friends of similar age are most likely
to share certain label types such as high school and college.  We find
that clipping the distributions makes little difference to accuracy while significantly
improving running time. However, the value of $K$ matters
significantly, and we detail these effects next.

\mypara{Recall of \OurModel}
Figure~\ref{fig:liftWRTbaseline} shows recall as a function of varying
number of friends $K$, against a baseline of \OurModel with $K=20$.
We observe that recall increases up to a certain $K$ and then
decreases --- $K=100$ for recall at $1$, and $K=200$ for recall at
$3$.  This demonstrates both the importance and the limits of
scalability: increasing the number of friends enables better inference
but beyond a point, more friends increase noise.
Thus, $K=100$ friends appear to be enough for
inference under \OurModel.

Figure~\ref{fig:liftWRTbaseline} also shows an increasing trend from
hometown to employer in the degree of improvement obtained over the
$K=20$ baseline. This is because (a) the baseline itself is best for
hometown and worst for employer, but also because (b) Facebook users
appear to have many more friends from label types other than from
their current employer.  The effect of this latter observation is that
if we only have a small $K$,
it is
very likely that the few friends from the same current employer are
not included in that limited set of friends (which we empirically verified).
As $K$ increases and such same-employer edges become available,
\OurModel can easily learn the reason for these edges (hence the
dramatic increase in recall), but label propagation will likely be
confused by the overall distribution of different employers among all
friends and therefore does not benefit equally from adding more
friends, as we show next.

\begin{figure*}[htbp]
\vspace{-0.5em}
\begin{center}
\begin{tabular}{cc}
\includegraphics[width=0.475\textwidth]{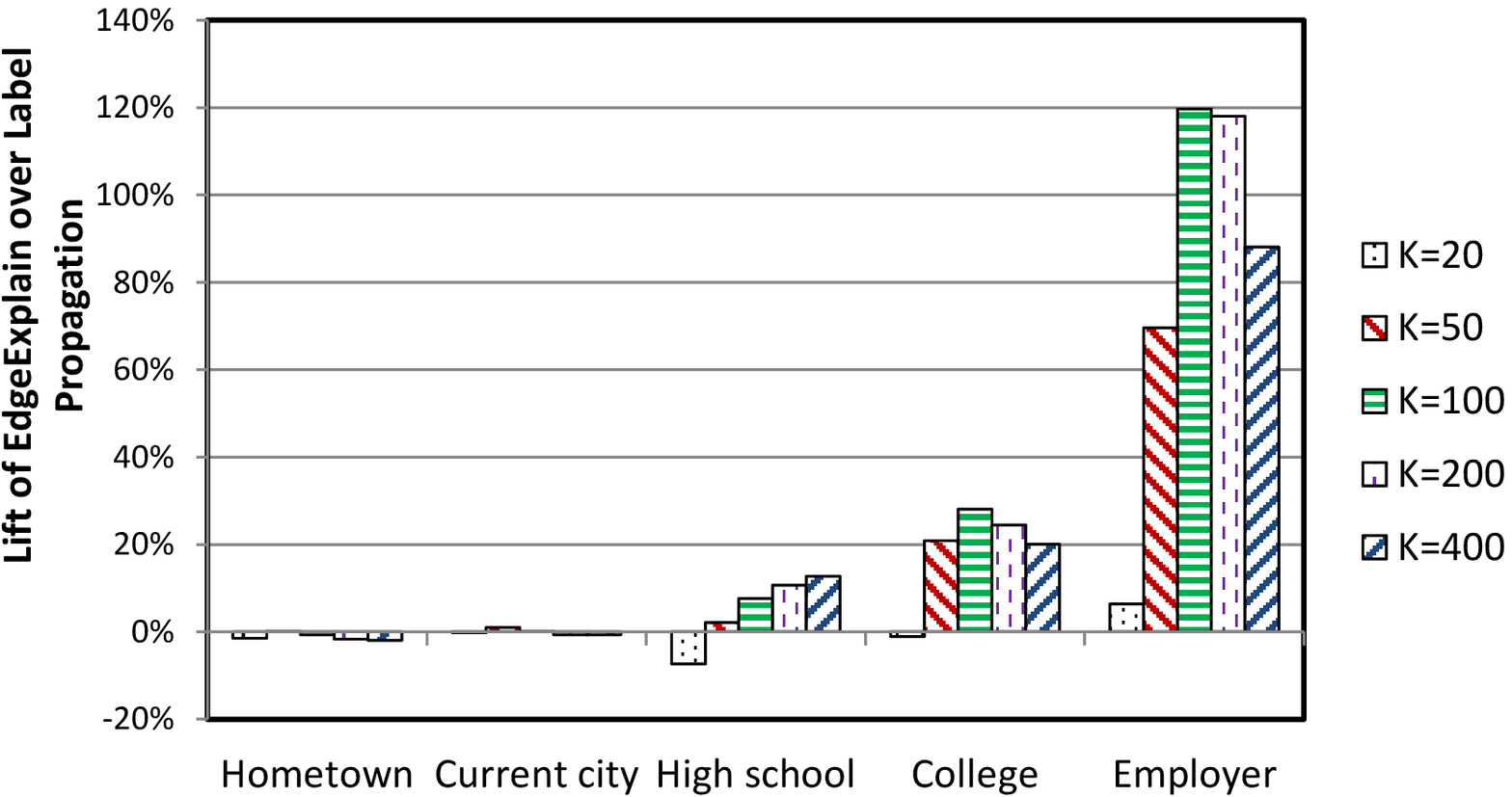} &
\includegraphics[width=0.475\textwidth]{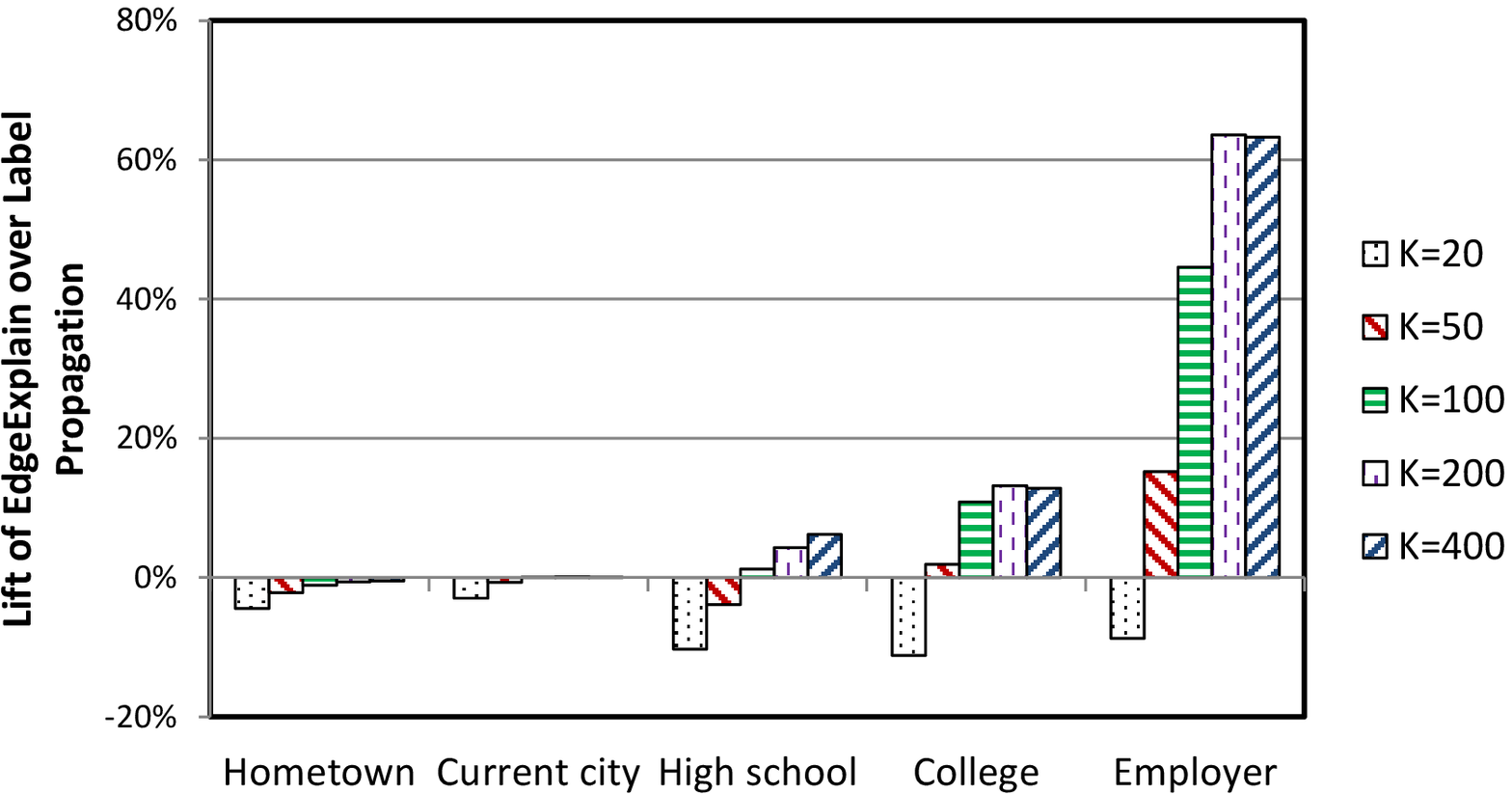}\\
(a) Recall at $1$ & (b) Recall at $3$
\end{tabular}
\caption{{\em Lift of \OurModel over Label Propagation:} Increasing
  the number of friends $K$ benefits \OurModel much more than label propagation
  for high school, college, and especially employer.}
\label{fig:liftWRTlp}
\end{center}
\vspace{-1em}
\end{figure*} 

\mypara{Comparison with Label Propagation}
Figure~\ref{fig:liftWRTlp} shows the lift in recall achieved by
\OurModel over Label Propagation as we increase $K$ for both.  We
observe similar performance of both methods for hometown and current
city, but increasing improvements for high school, college, and
employer.  This again points to the first two being easier to infer,
with the difficulty of inference increase with the latter label types.
With fewer employer-based friendships, the prototypical example of
Figure~\ref{fig:example} would also occur frequently, with Label
Propagation likely picking common employers of (say) hometown friends
instead of the less common friendships based on the actual
employer. By attempting to explain each friendship, \OurModel is able
to infer the employer even under such difficult circumstances, and the
ability to perform well even for under-represented label types makes
\OurModel particularly attractive.

\begin{table}[t]
\caption{{\em Lift in recall from using group memberships:} Inclusion
of group membership barely improves recall@3, even though it is an
orthogonal feature with wide coverage. Thus, information about label
types is already encoded in the network structure, and careful
modeling via \OurModel is sufficient to extract it.}
\label{tbl:groupMembership}
\begin{center}
\begin{small}
\begin{sc}
\begin{tabular}{lrr}
\hline
\abovespace\belowspace
Label Type & Recall at $1$ & Recall at $3$ \\
\hline
Hometown & $-0.1\%$ & $0.7\%$ \\
Current city &  $0.4\%$ & $1.0\%$ \\
High school  &  $0.1\%$ & $0.8\%$ \\
College & $-0.6\%$ & $1.0\%$ \\
Employer     &   $-2.8\%$ & $1.2\%$ \\
\hline
\end{tabular}
\end{sc}
\end{small}
\end{center}
\vspace{-1em}
\end{table}

\mypara{Inclusion of extra features}
In Section~\ref{sec:general}, we discussed how extra features could be
used within the \OurModel framework. In particular, we showed how the
fact that some users are members of groups can be used to infer (say)
their college, if the group turns out to be college-specific group.
Group memberships are extensive and provide information
that is orthogonal to friendships; thus, {\em a priori}, one would
expect the addition of group membership features to have significant
impact on label inference.

Table~\ref{tbl:groupMembership} shows the lift in recall for \OurModel
when group memberships are used in addition to $K=100$ friends. 
While the addition of group memberships increases the size of the
graph by $\approx 25\%$, the observed benefits for recall are minor:
a maximum lift of only $1.2\%$ for employer inference, and indeed
reduced recall at $1$ for several label types. 
Note that the lift in recall would have appeared very significant had we
compared it to Label Propagation with $K=100$; however, this gain
largely disappears when the friendships are considered in the
framework of \OurModel.
Thus, it is not merely the scalability of \OurModel, but also the
careful modeling of properties \Pone-\Pthree that makes group
membership redundant. 

Given the {\em a priori} expectations of the impact of group
memberships, this surprising result suggests that information
regarding our label types are already encoded in the structure of the
social network and hence the orthogonal information from the group
memberships actually turn out to be redundant.

\begin{figure}[ht]
\begin{center}
\centerline{\includegraphics[width=0.95\columnwidth]{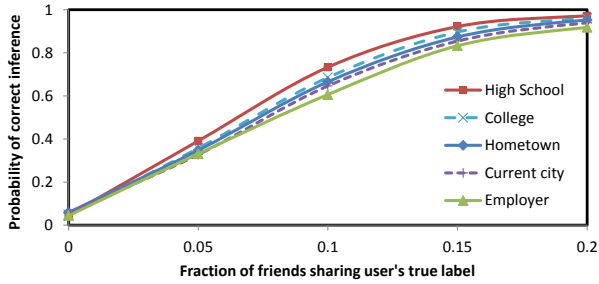}}
\caption{{\em Probability of correctly inferring (in the top-3) the
    value of a given label type $t$ for a user, given the fraction of friends with
	known label for $t$ who actually share the user's label for $t$:}
    All label types are
  broadly similar, with a fraction of $0.1$ usually being sufficient
  for inference.  For fraction $>0.2$, the plot flattens out.}
\label{fig:probCorrect}
\end{center}
\vspace{-2em}
\end{figure} 

\mypara{The limits of resolution}
Our model theoretically should be able to handle any number of label
types, but empirically this may not hold true for our network. 
How many friends sharing a certain label type (say, the same college)
does a user
need to have in order to correctly infer the
value of that label type?  To answer this, we select, for each user, the set of
friends whose label for the given label type $t$ is known, and we compute
the fraction that actually shared the user's label for $t$.
Figure~\ref{fig:probCorrect} shows the probability that
\OurModel correctly infers the user's label as a function of this
fraction (i.e., the correct label is among the top 3 predictions).
All label types are similar, though high school is somewhat easier and
employer harder; having $10-15\%$ of friends sharing a user's label is
sufficient to infer the label in our graph.
Note that certain label types are more
likely to be publicly declared than others, and this explains
differences in recall observed earlier.

\begin{figure}[ht]
\begin{center}
\centerline{\includegraphics[width=0.95\columnwidth]{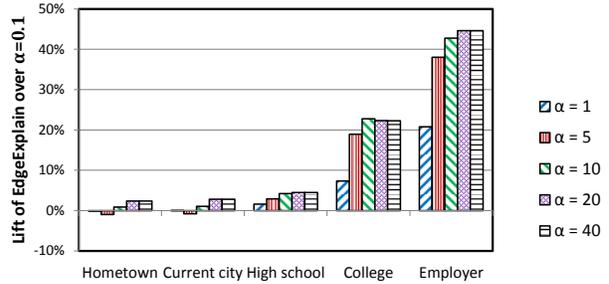}}
\caption{{\em Effect of $\alpha$:} Lift in recall at $1$ is plotted
for different values of $\alpha$, with respect to $\alpha=0.1$. i
The best results are for $\alpha\in[10,40]$.}
\label{fig:effectAlpha}
\end{center}
\vspace{-2em}
\end{figure} 

\mypara{Effect of $\alpha$}
Figure~\ref{fig:effectAlpha} shows that the lift in recall at $1$
for various values of the parameter $\alpha$, with respect to
$\alpha=0.1$.
Performance generally improves with increasing $\alpha$.  
Results for recall at $3$ are qualitatively similar, though the effect
is more muted.
We
find that $\alpha\in[10, 40]$ offer the best results, and \OurModel is
robust to the specific choice of $\alpha$ within this range.
Recall that with large $\alpha$, a single matching label is enough to
explain an edge, while with small $\alpha$, multiple matching labels
may be needed. Thus, the outperformance of large $\alpha$ provides
strong empirical validation of property \Ptwo (on our network).

\begin{figure}[ht]
\begin{center}
\centerline{\includegraphics[width=0.95\columnwidth]{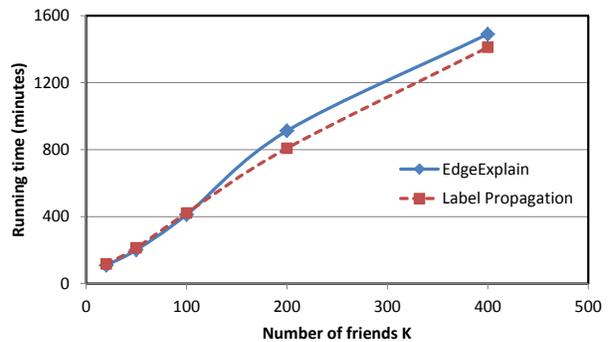}}
\caption{{\em Running time increases linearly with $K$.}}
\label{fig:runningTime}
\end{center}
\vspace{-2em}
\end{figure} 

\mypara{Running time}
Figure~\ref{fig:runningTime} shows the wall-clock time as a function
of $K$. The running time should depend linearly on the graph size,
which grows almost linearly with $K$; as expected, the plot is linear,
with deviations due to garbage collection stoppages in Java.


\section{Conclusions}
\label{sec:conc}

%

We proposed the problem of jointly inferring
multiple correlated label types in a large network and described the
problems with existing single-label models.  We noted that one
particular failure mode of existing methods in our problem setting is
that edges are often created for a reason associated with a particular
label type (e.g., in a social network, two users may link because they
went to the same high school, but they did not go to the same
college).  We identified three network properties that model this
phenomenon: edges are created for a reason \Pone, they are generally
created only for one reason \Ptwo, and sharing the same value for a
label type is necessary but not sufficient for having an edge between
two nodes \Pthree.

We introduced \OurModel, which carefully models these properties.
It leverages a gradient-based method for collective inference
which allows for fast iterative inference that is equivalent in
running time to basic label propagation.  Our experiments with a
billion-node subset of the Facebook graph amply demonstrate the
benefits of \OurModel, with significant improvements across a set of
different label types.  Our further analysis validates many of the
properties and intuitions we had about modeling networks, primarily
that one can achieve significant improvements if one considers and
models the {\em reason} an edge exists.  Whether one is interested in
inferring one or multiple label types, modeling these explanations
will have significant impact on the accuracy of the final predictions.

\bibliography{paper,stano,sofus}
\bibliographystyle{icml2014}

\end{document}